\documentclass[pdflatex,sn-mathphys-num]{sn-jnl}

\usepackage{graphicx}%
\usepackage{multirow}%
\usepackage{amsmath,amssymb,amsfonts}%
\usepackage{amsthm}%
\usepackage{mathrsfs}%
\usepackage[title]{appendix}%
\usepackage{xcolor}%
\usepackage{textcomp}%
\usepackage{manyfoot}%
\usepackage{booktabs}%
\usepackage{algorithm}%
\usepackage{algorithmicx}%
\usepackage{algpseudocode}%
\usepackage{listings}%

\usepackage{graphicx}
\usepackage{multicol}
\usepackage{mhchem,lipsum}
\usepackage{amsmath}
\usepackage{lineno}
\usepackage[utf8]{inputenc}
\usepackage{csquotes}

\usepackage{amsfonts}

\usepackage{tikz}
\usetikzlibrary{arrows,backgrounds}
\usepackage[utf8]{inputenc}
\usepackage{subfig}
\usepackage{mhchem,lipsum}
\usepackage{placeins}
\usepackage{adjustbox,color, colortbl}
\usepackage{booktabs}
\usepackage{multirow}
\usepackage{csquotes}

\usepackage{hyperref}       
\usepackage{url}            
\usepackage{booktabs}       
\usepackage{amsfonts}       
\usepackage{nicefrac}       
\usepackage{microtype}      
\usepackage{lipsum}
\usepackage{fancyhdr}       
\usepackage{graphicx}       
\graphicspath{{media/}}     
\usepackage{ dsfont }
\usepackage{amssymb}
\usepackage{algorithm}
\usepackage{dsfont}
\usepackage{amsmath}
\newcommand{\norm}[1]{\left\lVert#1\right\rVert}
\usepackage{ wasysym }
\usepackage{algorithm}
\usepackage{algpseudocode}

\theoremstyle{thmstyleone}%
\theoremstyle{thmstyletwo}%

\theoremstyle{thmstylethree}%

\begin{document}

\title[Article Title]{Gradient Enhanced Self-Training Physics-Informed Neural Network (gST-PINN) for Solving Nonlinear Partial Differential Equations}

\author*[1]{\fnm{Narayan} \sur{S Iyer}}\email{narayansiyer7@gmail.com}
\author*[2]{\fnm{Bivas} \sur{Bhaumik}}\email{bhaumikbivas@gmail.com}
\author[3]{\fnm{Ram} \sur{S Iyer}}\email{ramsiyer77@gmail.com}
\author[4]{\fnm{Satyasaran} \sur{Changdar}}\email{sach@di.ku.dk}
\affil[1]{\orgdiv{Department of Physics and Astronomy}, \orgname{National Institute of Technology Rourkela}, \orgaddress{\postcode{769008}, \state{Odisha}, \country{India}}}
\affil[2]{\orgdiv{Department of Mathematics}, \orgname{National Institute of Technology Rourkela}, \orgaddress{\postcode{769008}, \state{Odisha}, \country{India}}}
\affil[3]{\orgdiv{Department of Electronics Engineering}, \orgname{Rajiv Gandhi Institute of Petroleum Technology}, \orgaddress{\postcode{229304}, \state{Uttar Pradesh}, \country{India}}}
\affil[4]{\orgdiv{Department of Food Science}, \orgname{University of Copenhagen}, \orgaddress{\city{Copenhagen}, \country{Denmark}}}

\abstract{Partial differential equations (PDEs) provide a mathematical foundation for simulating and understanding intricate behaviors in both physical sciences and engineering.
With the growing capabilities of deep learning, data-driven approaches like Physics-Informed Neural Networks (PINNs) have been developed, offering a mesh-free, analytic type framework for efficiently solving PDEs across a wide range of applications. However, traditional PINNs often struggle with challenges such as limited precision, slow training dynamics, lack of labeled data availability, and inadequate handling of multi-physics interactions.
To overcome these challenging issues of PINNs, we proposed a Gradient Enhanced Self-Training PINN (gST-PINN) method that specifically introduces a gradient based pseudo point self-learning algorithm for solving PDEs. We tested the proposed method on three different types of PDE problems from various fields, each representing distinct scenarios. The effectiveness of the proposed method is evident, as the PINN approach for solving the Burgers’ equation attains a mean square error ($MSE$) on the order of $10^{-3}$, while the diffusion–sorption equation achieves an MSE on the order of $10^{-4}$ after $12,500$ iterations, with no further improvement as the iterations increase
In contrast, the MSE for both PDEs in the gST-PINN model continues to decrease, demonstrating better generalization and reaching an MSE on the order of $10^{-5}$ after 18,500 iterations.
Furthermore, the results show that the proposed purely semi-supervised gST-PINN consistently outperforms the standard PINN method in all cases, even when solution of the PDEs are unavailable. It generalizes both PINN and Gradient-enhanced PINN (gPINN), and can be effectively applied in scenarios prone to low accuracy and convergence issues, particularly in the absence of labeled data.}


\keywords{Physics Informed Neural Network, Non-Linear PDEs, Gradient Enhanced Pseudo Points, Gradient Residual Filtering, Self Semi-supervised Training}

\maketitle

\section{Introduction}
\subsection{Scientific And Physics Informed Machine Learning For Solving Non Linear PDEs}
Computational efficiency of traditional numerical methods in addressing partial differential equations (PDEs) arising in non trivial situations such as Riemannian manifolds, complex manifolds (i.e. specifically non Euclidean spaces), high dimensional spaces and inverse problem solving have always been a topic of concern. The advent of deep learning has facilitated data-driven models in PDE solving, yet their precision remains inherently tied to the size, distribution, and representativeness of the training data. Physics informed machine learning using physics informed neural networks \cite{raissi2019physics,cai2021physics,toscano2025pinns,wang2023expert,karniadakis2021physics,meng2025physics} has been successful in tacking the issue till an extent, which unlike traditional numerical solvers and data driven methods solves the PDE by encoding the physics information such as the PDE itself, boundary conditions, initial conditions, any other constraints such as symmetries etc. into the loss function, thereby converting numerical analysis problem into an optimization problem. The accuracy of PINNs over other traditional solvers can be attributed to mainly two of its charecteristics: firstly they are mesh free, due to which they can randomly sample points in the domain, and secondly they ideally do not require any labeled data for functioning, nevertheless a small amount of labeled data can help in better convergence to the solution. Despite theoretical guarantees, the accuracy and efficiency of ordinary PINNs remain insufficient for practical applications \cite{farea2024understanding,klapa2024machine}. A sparse amount of work targeted in this direction have been conducted in recent years. Numerous modifications to standard PINNs have been researched, yielding promising results in some cases \cite{changdar2024integrating}. However, robust models that consistently perform well across a broad range of PDE types with diverse applications in science and engineering remain elusive.
\subsection{Self Training}
Self training approaches \cite{amini2025self,triguero2015self,zou2019confidence}, which is a branch of semi-supervised learning \cite{hady2013semi,van2020survey} have become increasingly popular in recent years due to their ability in harnessing limited labeled data and plentiful unlabeled datasets, thereby combining them to enhance predictive performance across various tasks. The core concept of self training revolves around repeatedly assigning pseudo labels to unlabeled samples having confidence scores exceeding a specified threshold, augmenting the labeled dataset and retraining. Some of the popular pseudo-labeling strategies include threshold based methods, proportion based methods, curriculum learning based methods majority vote classifiers and adaptive thresholding methods. A mathematical description of self training is provided in \cite{amini2025self}.

\subsection{Related works}
To resolve the challenges related to PINNs (in terms of accuracy, efficiency, solving PDEs in complex manifolds, solving higher dimensional PDEs, etc.), various modifications to the standard PINN framework have been proposed by several researchers, thereby yielding novel algorithms and models. For instance, the Bayesian physics-informed neural network (B-PINNs) was proposed by L Yang et al. \cite{yang2021b} which integrates Bayesian neural networks with traditional PINNs to solve forward and inverse PDEs with noisy measurements as input data, which enabled uncertainty quantification and more accurate predictions compared to PINNs, especially in extremely noisy scenarios. A similar augmentation to the standard PINN framework was formulated by A D Jagtap et al. \cite{jagtap2020conservative}, the conservative physics-informed neural network (cPINN), which functions by decomposing the computational domain into discrete sub-domains, each with its own deep or shallow neural network (which is dependent on the complexity of the solution in a particular sub-domain) trained separately and by enforcing flux continuity at the sub-domain interfaces for preserving the conservation laws. cPINN has been analyzed to have multiple advantages such as reduced spread of error across the domain, efficient fine-tuning of hyper parameters and enhanced representation ability of the network. Albeit exceptionally accurate across a wide range of PDEs, cPINN fails to perform in situations involving complex-geometrical and highly irregular domains. To address this shortcoming of cPINN, G E Karniadakis et al. presented the Extended physics-informed neural network (XPINN) \cite{jagtap2020extended}, which is based on a generalized space-time computational domain decomposition algorithm for PINNs. In addition to preserving the advantages of cPINN over the classical PINNs, XPINN was proved to be highly applicable to multi-physics and dynamic interface problems, and also hold certain exclusive benefits while addressing the same. Moreover, the domain decomposition algorithm based on XPINN is extendable to any type of PDEs, regardless of its physical nature. Another non-trivial limitation of PINNs which have been less researched on is the Unbalanced Prediction Problem (UPP) which primarily arises due to the equal treatment of easy (where solutions are relatively smooth) and hard (points in regions with steep gradients, points near boundaries, etc. where approximation of solution is relatively harder) sample points by PINNs, thereby leading to unstable learning of the network. To overcome this problem, S Duan et al. proposed cognitive physics-informed neural network (CoPINN) \cite{duancopinn}, which operates by implicitly evaluating the hardness of every sample during training, based on the PDE residual gradient. Apart from these proposals, numerous modifications on the standard PINN structure have been researched on, tailored at specifically finding applications in fluid dynamics \cite{zhao2024comprehensive,choi2022physics}, power systems \cite{stiasny2024pinnsim} and financial mathematics \cite{nuugulu2024physics,bai2022application}.

\subsection{Overview of the paper and our contribution}
The article is organized as follows: In section (2), a concise and mathematically streamlined overview of PINN is presented followed by the gPINN methodology for solving non-linear PDEs, concluding with the discussion on formulation of the total loss function for the gPINN model. In section (3), we introduce our proposed gST-PINN model. Subsequently, in the subsections (3.1) and (3.2), the overview of the framework, the gST-PINN model architecture and the gradient enhanced pseudo point generation algorithm is discussed in detail. In section (4), various experimental results (section (3.1): Burgers' equation, section (3.2): diffusion reaction equation, section (3.3): diffusion sorption equation) have been demonstrated to support the arguments made in section (3). Consequently, in section (3.4), a detailed comparison of the gST-PINN model's performance with the other state-of-the-art models have been discussed. In section (5) (appendix section), a mathematical overview of semi-supervised learning and self training is provided. Finally, the concluding remarks and potential future research trajectories have been summarized in section (6).\\
Our key contributions include:
\begin{enumerate}
    \item We formulate a novel gradient enhanced pseudo point generation algorithm which utilizes the gradient information of the PDE residual under consideration for sorting and selecting the collocation points for pseudo labeling.
    \item We propose a novel gradient enhanced self training physics informed neural network model (gST-PINN) based on our novel pseudo labeling algorithm.
    \item We provide an evaluation of the accuracy our model in terms of the $MSE$ and $L_2$ relative errors of the predicted solution to various non-linear PDEs and perform a comparative analysis with a wide range of scientific machine learning settings.
\end{enumerate}

\section{Methodology}
\subsection{Physics Informed Neural Network (\text{PINN}) for solving PDEs} \label{sec_2.1_PINN_for_PDES}
We first consider the general form of a PDE for the solution $u(\textbf{x})$ defined on a domain $\Omega \subset \mathds{R}^N$ with a parametrization $\zeta = (\zeta_1,~ \zeta_2,...)$ and $\textbf{x}=(x_1,~x_2,...,~x_N) \in \Omega:$\\
\begin{equation} \label{PDE-residual}
    G(\textbf{x}; \frac{\partial u}{\partial x_1},..., \frac{\partial u}{\partial x_N}; \frac{\partial ^{2} u}{\partial x_1 \partial x_1},...,\frac{\partial ^2 u}{\partial x_1 \partial x_N},...,\frac{\partial ^2 u}{\partial x_N \partial x_N},...;\zeta ) =0
\end{equation}
along with the boundary condition (Dirichlet, Periodic, etc.) defined on $\partial \Omega$ and additional constraints (such as initial condition, symmetries, etc.) respectively:
\begin{equation} \label{bound_ini}
    D(u;\textbf{x}) = 0 ~~~~\mathrm{and} ~~~~S(u,\textbf{x})=0.
\end{equation}
The PDE can be solved by constructing a neural network $\hat{u}(\textbf{x} , \tilde{\theta } )$ using trainable parameters $\tilde{\theta} $ to estimate the solution $u(\textbf{x})$ and consecutively optimizing $\tilde{\theta }$ using the constraints dictated by the PDE (through equations (\ref{PDE-residual}) and (\ref{bound_ini})). A set of clustered points $\mathcal{T} = \mathcal{T}_{G} \cup \mathcal{T}_{D} \cup \mathcal{T}_{S} = \{\textbf{x}_1, \textbf{x}_2,...,\textbf{x}_{|\mathcal{T}|}\}$ of size $|\mathcal{T}|$ represents the collocation points chosen randomly to train the network where $\mathcal{T}_G \subset \Omega$ of size $|\mathcal{T}_{G} |$, $\mathcal{T}_{D} \subset \partial \Omega$ of size $|\mathcal{T}_{D}|$ and $\mathcal{T}_{S} \subset \Omega$ of size $|\mathcal{T}_{S} |$ separately denotes the collocation points on the domain, boundary and for additional constraints respectively. To optimize the parameters $\tilde{\theta}$, we utilize the loss function $\mathcal{L} (\tilde{\theta} , \mathcal{T} )$ defined as
\begin{equation} \label{loss_without_grad}
\mathcal{L}(\tilde{\theta};\mathcal{T}) = w_G\mathcal{L}_G(\tilde{\theta};\mathcal{T}_G)+ w_D\mathcal{L}_D(\tilde{\theta};\mathcal{T}_D) + w_S\mathcal{L}_S(\tilde{\theta};\mathcal{T}_S)
\end{equation}
where $w_G$, $w_D$, $w_S$ are the weights and the individual losses $\mathcal{L}_G$, $\mathcal{L}_D$ and $\mathcal{L}_D$ in accordance with equations (1) and (2) can be defined respectively as 
\begin{equation} \label{res_loss}
\mathcal{L}_G(\tilde{\theta};\mathcal{T}_G)= \frac{1}{|\mathcal{T}_G|} \sum _{\textbf{x} \in \mathcal{T}_G} \norm{G(\textbf{x};\frac{\partial \hat{u}}{\partial x_1},...;\frac{\partial^2 \hat{u}}{\partial x_1 \partial x_1},...;\zeta)}_2^2
\end{equation}
and
\begin{equation} \label{bound_loss}
    \mathcal{L}_D(\tilde{\theta};\mathcal{T}_D)= \frac{1}{|\mathcal{T}_D|}  \sum_{x\in \mathcal{T}_D}\norm{\mathcal{D}(\hat{u},\textbf{x})}_2^2
\end{equation}
and
\begin{equation} \label{ini_sym_loss}
    \mathcal{L}_S(\tilde{\theta};\mathcal{T}_S)= \frac{1}{|\mathcal{T}_S|}  \sum_{x\in \mathcal{T}_S}\norm{\mathcal{S}(\hat{u},\textbf{x})}_2^2
\end{equation}
Here, $\norm{.}_2$ indicates the $L^2$ norm. The optimal $\tilde{\theta}$ upon training of the network $u(\textbf{x},\tilde{\theta} )$ using the above loss function is $\tilde{\theta} ^* = \mathrm{arg ~min}_{\tilde{\theta} } \mathcal{L}(\tilde{\theta} ;\mathcal{T}).$ Alternatively, the value of function $G$ in equation (1) is called the PDE residual and $\mathcal{T}_G$ the set of residual points. 

\subsection{Gradient enhanced physics informed neural network (\text{gPINN}) for solving PDEs} \label{sec_2.2-gPINN_for_PDE}
gPINN \cite{yu2022gradient} improves the traditional PINN framework by leveraging the derivative information of the PDE residual and using it to enrich the loss function. Referring to the PDE given in equation (1) again and considering the fact that $G$ vanishes for all $\textbf{x}\in \Omega$, a direct inference can be made that the partial derivative of $G$ with respect to each component of $x$ (i.e $x_1,x_2,...,x_N$) is also zero for all $x\in \Omega$.
\begin{equation} \label{grad_residual}
\therefore \nabla G(\textbf{x}) = \left(\frac{\partial G}{\partial x_1}, \frac{\partial G}{\partial x_2},...,\frac{\partial G}{\partial x_N}\right)=\textbf{0}, ~~~\forall \textbf{x}\in \Omega.
\end{equation}
The additional loss term as a consequence of the above information is 
\begin{equation}
     \mathcal{L}_{g_{i}}(\tilde{\theta} ;\mathcal{T}_{g_i} ) = \frac{1}{|\mathcal{T}_{g_i}|} \sum_{x\in \mathcal{T}_{g_i}} \norm{\frac{\partial G}{\partial x_i}}_2 ^2
\end{equation}
where $\mathcal{T}_{g_{i}}$ represents the set of collocation points for the partial derivative $\frac{\partial G}{\partial x_{i}}$, with size $|\mathcal{T}_{g_{i}} |$. Upon using appropriate weights $w_{g_{i}}$, the total loss function of equation (3) modifies to 
\begin{equation} \label{loss_with_grad}
     \mathcal{L} = w_G \mathcal{L}_G + w_D\mathcal{L}_D + w_S \mathcal{L}_S + \sum_{i=1}^N w_{g_i}\mathcal{L}_{g_i}(\tilde{\theta};\mathcal{T}_{g_i}).
\end{equation}

\section{Gradient Enhanced Self Training Physics Informed Neural Network (\text{gST-PINN}): Proposed Approach} \label{sec_3_gSTPINN}
This section outlines the proposed gST-PINN framework in detail, which implicitly combines gPINN framework with the pseudo point generation strategy, thus developing a novel gradient enhanced self training model that is physics aware. 
\subsection{Overview of the framework}
\begin{figure}[t!] 
	\centering
	\hspace*{-2.0cm} 
	\includegraphics[width=6.5in]{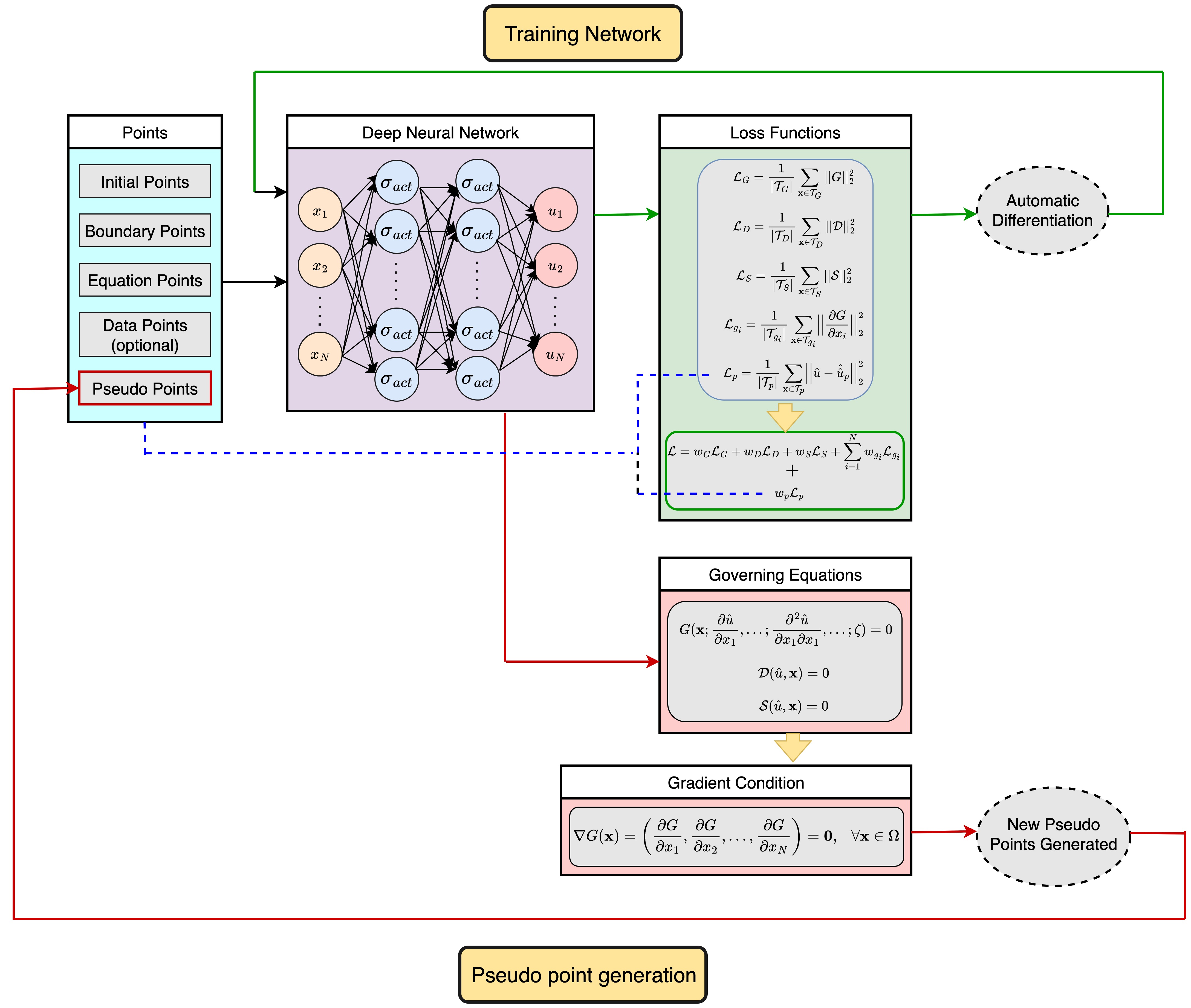}%
	\caption{Pipeline/Architecture of the gST-PINN model \textbf{(section (\ref{sec_3_gSTPINN}))}. The green line depicts the neural network training process fulfilled by minimizing the loss function given by equation (\ref{loss_with_grad}) \textbf{(section (\ref{sec_2.1_PINN_for_PDES}))} and the red line represents the gradient enhanced pseudo point generation algorithm/strategy accomplished by filtering and labeling the collocation points with considerably small PDE and gradient residuals given by equations (\ref{PDE-residual}) and (\ref{grad_residual}) \textbf{(section(\ref{sec_3.2_pse_gen_stra}), Algorithm \ref{algorithm_1}}. Both the processes, apart from the incorporation of the physics based information (PDE residual, initial and boundary conditions, additional constraints and symmetries, etc.), also utilizes the gradient details \textbf{(section (\ref{sec_2.2-gPINN_for_PDE}))} for escalating the overall accuracy of the model.}
	\label{Pipeline/Architecture of the gST-PINN model}
\end{figure}

The architecture of the proposed framework is illustrated in figure (\ref{Pipeline/Architecture of the gST-PINN model}). The network takes in $(x_1,~x_2,~x_3,...,~x_N) \in \mathds{R}^N$ (which in most cases will be coordinates in 3 dimensions constituting of the time coordinate $t$ and the regular spatial coordinates $x,~y$) as inputs and estimates $(v_1,~v_2,...,~v_M) \in \mathds{R}^M$ (which in most cases will be the vector fields, tensor fields and physical quantities such as velocity fields in two dimensions $(u,v)$, density $\rho$, etc.) as outputs. Each iteration comprises two key steps: network training and gradient enhanced pseudo label generation which are represented in green and red colored lines respectively in figure (\ref{Pipeline/Architecture of the gST-PINN model}).\\
1. $Network ~Training~:$ This part of the framework is similar to the normal PINNs. For the prediction of the solution $\hat{u}$ for the governing PDE, three sets of training points (or training data), namely residual points, boundary points and auxiliary constraint points such as collocation points for initial conditions, symmetry conditions, etc. are fed as inputs to the network as already mentioned in the above sections as $\mathcal{T}_G$, $\mathcal{T}_D$ and $\mathcal{T}_S$ respectively. The residual points are referred to as equation points in certain articles. The residual points are unsupervised points randomly sampled from the $N$-dimensional domain used to address the task of learning physics information through the minimization of residual loss (\ref{res_loss}) and thereby training the network by standard back propagation algorithm. The boundary points and the auxiliary constraint points can be either non-labeled or labeled data depending on whether the boundary and auxiliary conditions are given explicitly in analytic form (as in equation (\ref{bound_ini})) or not. If given in non-labelled form, the loss function can be modeled as given in equations (\ref{bound_loss}) and (\ref{ini_sym_loss}) to learn the corresponding information. Ideally, PINNs do not require any supervised data for learning the PDE solution. But the provision of few intra-domain labeled data can help in improving accuracy and efficiency in terms of convergence speed.\\
2. $Gradient ~Enhanced ~Pseudo ~Label~ Generation~:$ This part of the iteration is what distinguishes the proposed framework from ordinary PINN and gPINN, thereby improving its accuracy. We introduce a novel gradient-enhanced pseudo-label generation technique that incorporates pseudo-labels, generated based on the PDE residual and its gradient information, into the training process. In gST-PINN, following a certain number of iterations, the neural network generates predictions for all sample points and employs them to compute the governing equation's residual as well as partial derivative of residual with respect to all the $N$ coordinates $x_1,~x_2,...,~x_N$ (or in a compact way, the gradient of residual, i.e. $\nabla \equiv ( \frac{\partial} {\partial x_1},~\frac{\partial}{\partial x_2},...,\frac{\partial}{\partial x_N} ).$ For $(x_1,~x_2,~x_3) = (x,~y,~t)$, ~$\frac{\partial}{\partial x},~ \frac{\partial}{\partial y},~ \frac{\partial}{\partial t}$ are computed).

\subsection{Pseudo Point Generation Strategy} \label{sec_3.2_pse_gen_stra}
Generation of pseudo labels from selection followed by filtration of unsupervised data is a pivotal procedure in self training. However, the conventional pseudo-labeling method in classification, which relies on confidence thresholds to sort unlabeled data, cannot be directly applied to PINNs since solving a PDE is fundamentally different from solving a classification problem.

The pseudo point generation and labeling strategy of gST-PINN revolves around three crucial hyperparameters: q, the pseudo-label candidate rate or maximum rate, r, the pseudo-label selection threshold or the stable coefficient and p, the (pseudo-label) update frequency. The algorithm adopted by gST-PINN for the pseudo point selection and labeling is conspicuous through algorithm \ref{algorithm_1}. Primarily, upon the initialization of the pseudo point coordinates as empty sets, the network predicts solutions $\hat{u}$ (of the equation (\ref{PDE-residual}) with constraints given by equation (\ref{bound_ini})) corresponding to each of the $|\mathcal{T}_G|$ collocation points fed to it and the PDE residual denoted by $R_G$ based on the predictions are computed (lines 1-5). Now, the algorithm ranks the collocation points based on the residual and the top $q\%$ points with smallest residuals are sorted out for next filtration (line 6). This completes the initial phase of the algorithm.\\
The subsequent phase commences with the computation of the gradients of the PDE residual $G$ with respect to each of the $N$ coordinates $x_0$ to $x_{N-1}$, denoted by $g_i = \frac{\hat{\partial G}}{\partial x_i}$. Now, the residuals corresponding to each of the $N$ PDEs represented by equation (\ref{grad_residual}) for each of the initially sorted out top $q\%$ points in the first phase are evaluated. Furthermore, the gST-PINN generates $N$ rankings of the $|\mathcal{T}_G| \times q$ collocation points (termed as idx$_i$; i = 0 to $N-1$), each corresponding to the $N$ coordinates, on the basis of the respective gradient residual (equation (\ref{grad_residual})) and the top $q\%$ points among the available sample points with the smallest gradient residuals are promoted as candidates for pseudo points. This concludes the second phase of the algorithm (lines 7-13).\\
In the third phase, the algorithm keeps track of $N$ flag accumulators, each corresponding to the $N$ coordinates for the selection and labeling of pseudo points. The flag accumulators are updated (line 16) and if all the $N$ flag accumulators flag$_i$ to flag$_{i+N-1}$ exceeds the pseudo label selection threshold $r$ (line 18), the double filtered $|\mathcal{T}_G|\times q^2$ collocation points are elected as pseudo points. This summarizes the third phase of the gST-PINN algorithm (lines 14- 23).
In the last phase, the flag counters of the $|\mathcal{T}_G|(1-q^2)$ collocation points which fail to qualify as pseudo points are reset to 0, which is substantial in order to disqualify those candidate points with low residuals, but inconsistent, thereby leveraging the overall accuracy of the gST-PINN model (lines 24-28).   

\begin{algorithm} 
\caption{Pseudo point generation}
\begin{algorithmic}[1]
\State ${x_{0}}_p, {x_1}_p, {x_2}_p,...,{x_{N-1}}_p, u_p \gets \emptyset$
\For{$i = 0$ to $|\mathcal{T}_G| - 1$}
    \State $\hat{u}^i \gets \mathcal{N}^{(k)}(x_0^i, x_1^i,...,x_{N-1}^i)$
    \State $R_G^i \gets G(x_0^i,x_1^i,...,x_{N-1}^i,\frac{\partial \hat{u}^i}{\partial x_0^i}, \frac{\partial \hat{u}^i}{\partial x_1^i},...,\frac{\partial \hat{u}^i}{\partial x_{N-1}^i},\frac{\partial^2\hat{u}^i}{\partial {x_0^i}^2},\frac{\partial ^2\hat{u}^i}{\partial x_0^i \partial x_1^i},...,\frac{\partial^2\hat{u}^i}{\partial {x_{N-1}^2}^i},...,\hat{u}^i, \zeta)$
\EndFor
\State $\text{idx} = \text{argPartition}(R_G, x_0, x_1,...,x_{N-1}, \hat{u}, q)$
\For{$i = 0$ to $N-1$}
    \State $g_i = \hat{\frac{\partial G}{\partial x_i}}$
    \For{$j=0$ to $|\mathcal{T}_G| \times q -1$}
        \State ${R_g}_{i}^j \gets g_i(x_0^{idx[j]},x_1^{idx[j]},..,x_{N-1}^{idx[j]},\frac{\partial \hat{u}^{idx[j]}}{\partial x_0^{idx[j]}}, \frac{\partial \hat{u}^{idx[j]}}{\partial x_1^{idx[j]}},..,\frac{\partial \hat{u}^{idx[j]}}{\partial x_{N-1}^{idx[j]}},\frac{\partial^2\hat{u}^{idx[j]}}{\partial {x_0^{idx[j]}}^2},\frac{\partial ^2\hat{u}^{idx[j]}}{\partial x_0^{idx[j]} \partial x_1^{idx[j]}},..,\frac{\partial^2\hat{u}^{idx[j]}}{\partial {x_{N-1}^2}^{idx[j]}},..$
        \State $..,\hat{u}^{idx[j]}, \zeta)$
\EndFor
\State $\text{idx}_i = \text{argPartition}({R_g}_i, x_0, x_1,...,x_{N-1}, \hat{u}, q)$
\EndFor
\For{$i = 0$ to $N-1$}
    \For{$j = 0$ to $|\mathcal{T}_G| \times q^2 - 1$}
    \State $\text{flag}_i[\text{idx}_i[j]] \mathrel{+}= 1$
    \EndFor
    \If{$\text{flag}_i[\text{idx}_i[i]],~\text{flag}_{i+1}[\text{idx}_{i+1}[i]],...,~\text{flag}_{i+N -1}[\text{idx}_{i+N-1}[i]] > r$}
        \State ${x_0}_p \gets {x_0}_p \cup x_0^i$
        \State ${x_1}_p \gets {x_1}_p \cup x_1^i$ 
        \State $u_p \gets u_p \cup \hat{u}^i$
    \EndIf
\EndFor
\For{$i = 0$ to $N - 1$}
    \For{$j= |\mathcal{T}|_G \times q^2$ to $|\mathcal{T}_G| -1$}
    \State $\text{flag}_i[\text{idx}[j]] = 0$
\EndFor
\EndFor
\end{algorithmic} \label{algorithm_1}
\end{algorithm}

\section{Experimental Results}
To assess the performance of our proposed gST-PINN approach, we conduct a comprehensive set of experiments on various PDEs representing diverse fields and scenarios. A comparative performance analysis with the standard PINN method is also presented in Table 1.

\begin{figure}[h]
	\centering
	\hspace*{-2.2cm} 
	\includegraphics[width=6.5in]{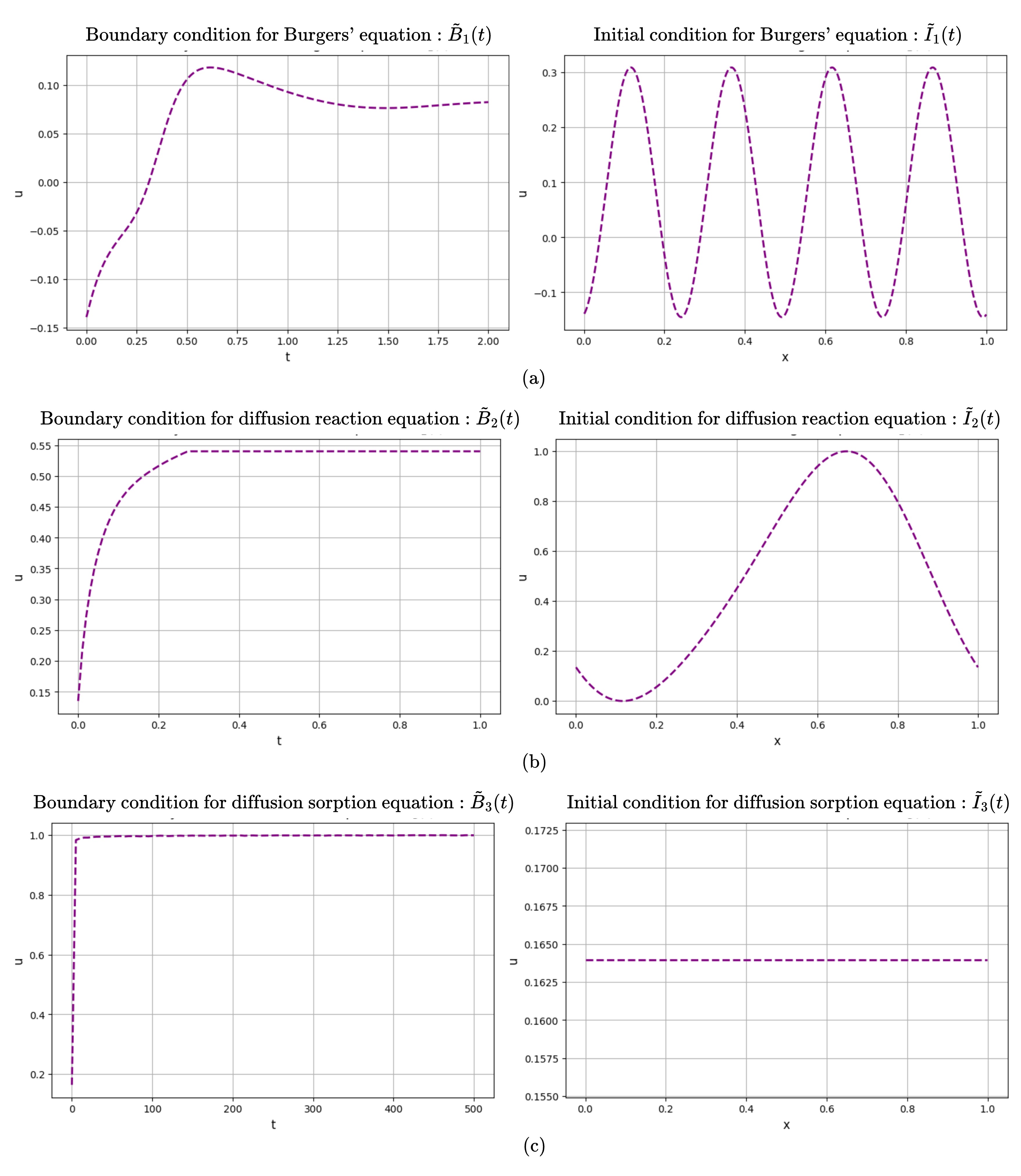}%
	\caption{Functional form of initial and boundary conditions of \textbf{(a)}: Burgers' equation \textbf{(section(\ref{sec_4.1_burgers'}))}, \textbf{(b)}: diffusion reaction equation \textbf{(section (\ref{sec_4.2_diff_react}))}, \textbf{(c)}: diffusion sorption equation \textbf{(section(\ref{sec_4.3_diff_sorp}))}.}
	\label{boundary_initial_all_equations}
\end{figure}

\begin{figure}[t!]
	\centering
	\hspace*{-3.4cm} 
	\includegraphics[width=7.0in]{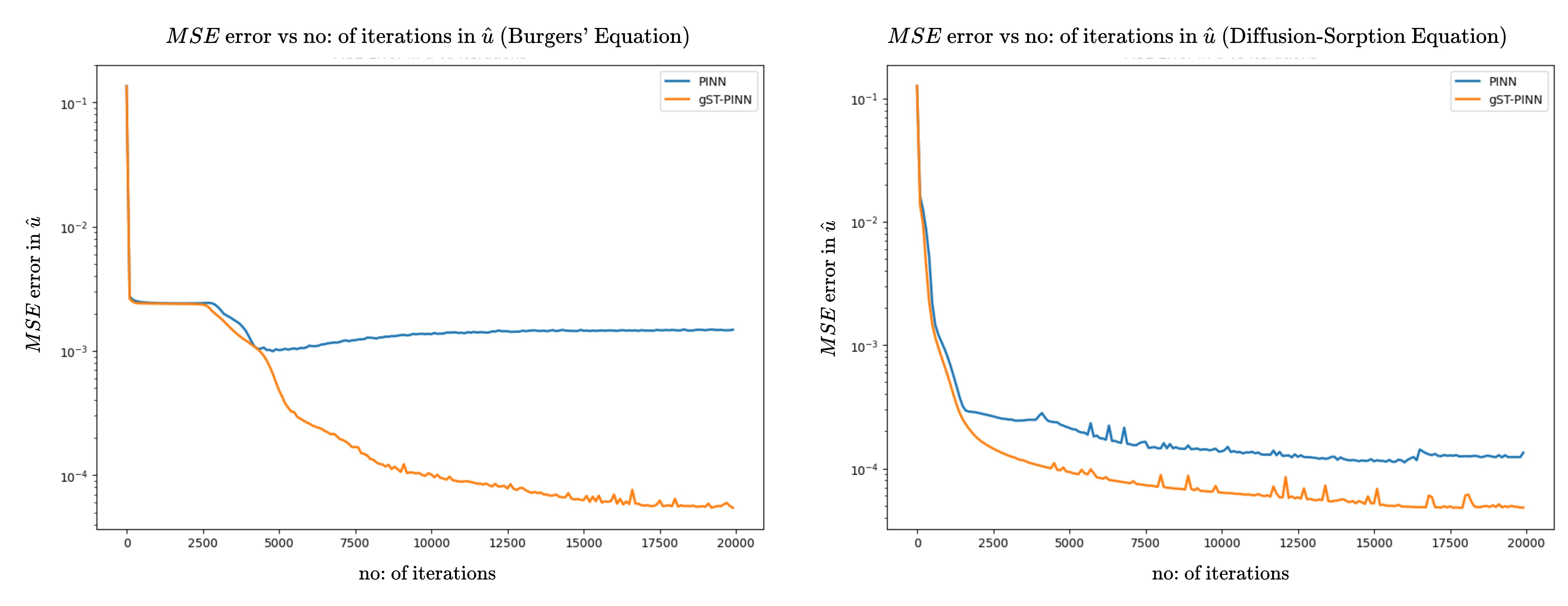}%
	\caption{Plots of $MSE$ error in $\hat{u}$ vs number of iterations in gST-PINN and PINN models for the approximation of solution to the \textbf{Burgers' equation} (\ref{sec_4.1_burgers'}) and \textbf{diffusion-sorption equation} (\ref{sec_4.3_diff_sorp}). It is conspicuous that the $MSE$ errors in PINN model converge very quickly (for example, approximately after 7500 iterations in case of Burgers' equation, eventually achieving the $\mathcal{O}(10^{-3})$ and 12500 iterations in case of diffusion-sorption equation, eventually achieving $\mathcal{O}(10^{-4})$), while the $MSE$ errors in gST-PINN model continue to decrease till the last iteration.}
	\label{MSE_errors_vs_iterations_plot}
\end{figure}

\subsection{Burgers Equation} \label{sec_4.1_burgers'}
\begin{figure}[h]
	\centering
	\hspace*{-1.5cm} 
	\includegraphics[width=6.5in]{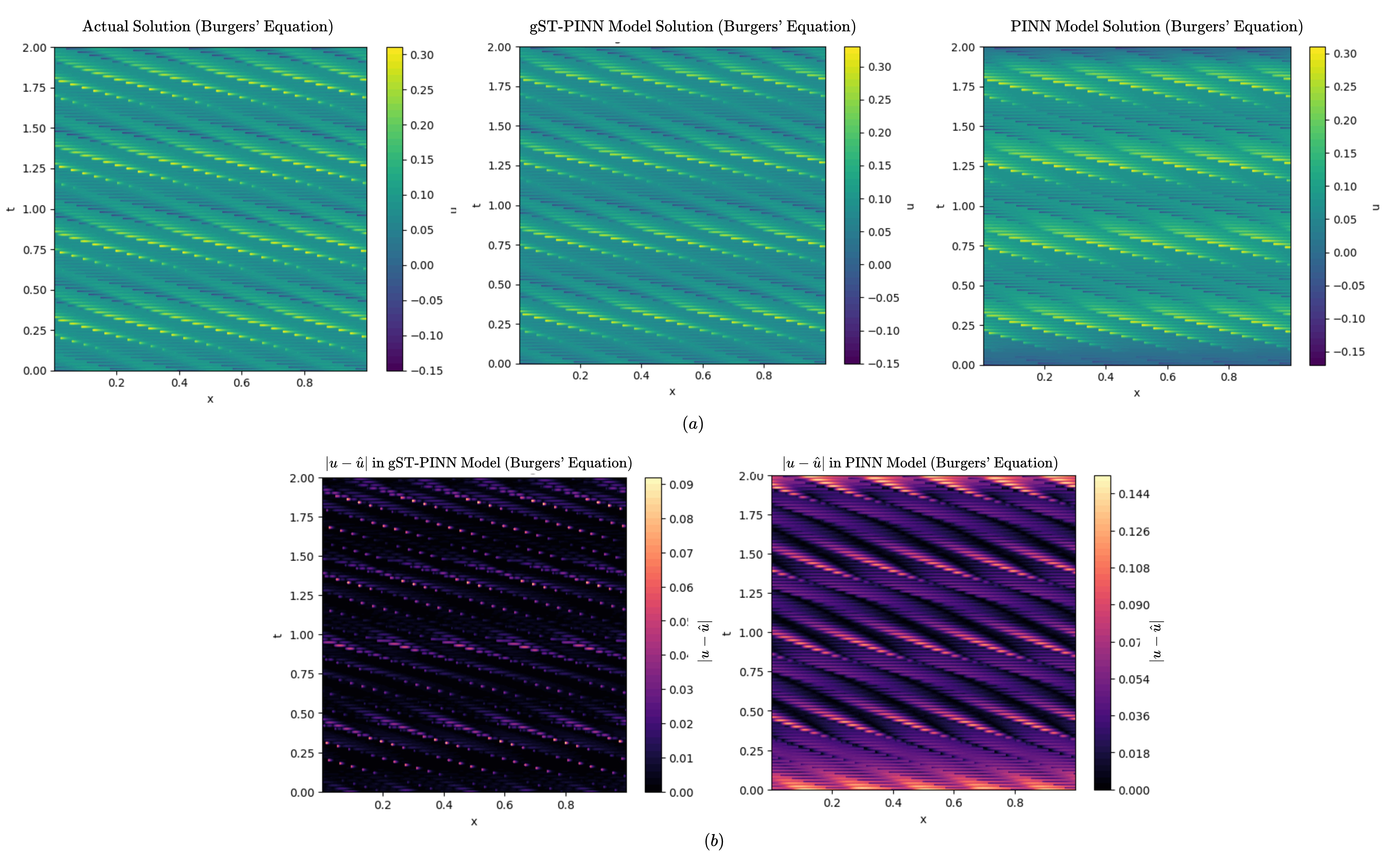}%
	\caption{\textbf{(Section (\ref{sec_4.1_burgers'}), Burgers' equation)} \textbf{(a)}: Comparison of the actual solution $u$ of Burgers' equation with its predicted solutions $\hat{u}$ using gST-PINN and PINN models. \textbf{(b)}: Comparison of the absolute error in $u$, i.e., $|u-\hat{u}|$ predicted by the gST-PINN and PINN model. From both \textbf{(a)} and \textbf{(b)}, it can be clearly observed that the prediction $\hat{u}$ of the solution $u$ for Burgers' equation performed by the gST-PINN model is more accurate in comparison with the PINN model, i.e., $|u(x)-\hat{u}(x)|_{gST-PINN} \leq |u(x)-\hat{u}(x)|_{PINN}$, $~~\forall x \in \Omega_{Burgers'}$, where $\Omega_{Burgers'}$ denotes the domain on which the Burgers' equation is defined.}
	\label{Burgers_equation_plot_diagram_1}
\end{figure}

\begin{figure}[h]
	\centering
	\hspace*{-2.2cm} 
	\includegraphics[width=6.5in]{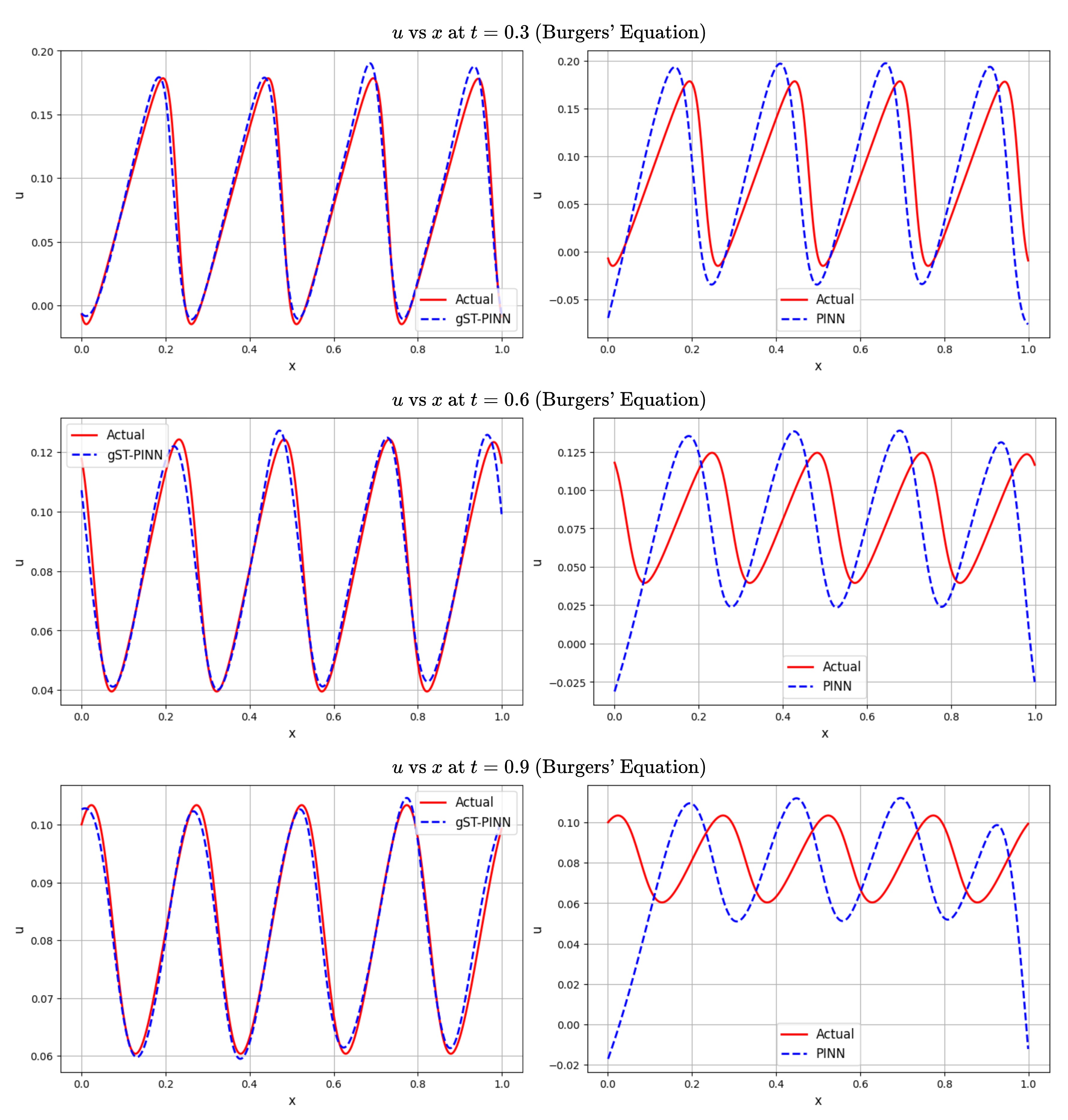}%
	\caption{\textbf{(Section (\ref{sec_4.1_burgers'}), Burgers' equation)} Performance comparison analysis of gST-PINN model and the normal PINN model for solving the Burgers' equation. The analysis has been done by studying the "$\textbf{u}$ vs $\textbf{x}$" plots at three different time instances, $\textbf{t=0.3,~0.6,~0.9}$. It has been examined from the plots that, initially \textbf{(t=0.3)}, both the models exhibit roughly identical accuracy, but at later intervals \textbf{(t=0.6,~0.9)}, the accuracy of the PINN model considerably deteriorates whereas the gST-PINN model succeeds in preserving the accuracy level exhibited initially, thereby proving to be more consistent compared to the PINN model.}
	\label{Burgers_equation_diagram_2}
\end{figure}

For a field $u(x,t)$, the general form of Burgers equation (also known as viscous Burgers equation or Bateman-Burgers equation) \cite{ortiz2024solving,li2023assessing} is given as
\begin{equation}
    \frac{\partial}{\partial t}(u(t,x)) + \frac{\partial}{\partial x}(u^{2}(t,x)/2) = (\eta_v / \pi) \frac{\partial ^{2}}{\partial x^{2}}, (u(t,x))~~~~~x\in(0,1),~t\in[0,2].
    \label{burg_pde}
\end{equation}
Here, $\eta_v$ (\textit{kinematic viscosity}) $\rightarrow$ 0.01.\\
The initial condition for the above given Burgers equation (\ref{burg_pde}), represented as a linear combination of sinusoidal functions is provided in equation (\ref{burg_ini}) as 
\begin{equation} \label{burg_ini}
    u(0,x)=u_{0}(x) = \sum_{i=1}^{N} \lambda_{i} \sin (2\pi f_{i}/L_x)x + \Psi_{i} \approx \tilde{I}_{1}(x) ~~~~~x\in (0,L_{x})
\end{equation} 
$\lambda_{i}$ (\textit{amplitude of the $i^{th}$ sinusoidal function}) $\in$ [0,1]\\
$\Psi_i$ (\textit{phase shift of $i^{th}$ sinusoidal function} $\in$ (0,2$\pi$)\\
$f_i$ (\textit{mode number determining the number of oscillations over the domain}) $\in$ $\mathds{N} := \{1,2,3,...\}$\\
$L_x$ (\textit{domain length in the spatial direction x}) $\rightarrow$ 1\\
The boundary condition considered is the periodic boundary condition, i.e.
\begin{equation} \label{burg_bound_cond}
u(t,0)=u(t,1)\approx \tilde{B}_{1}(t) ~\forall t\in[0,2]
\end{equation}
The plots of $\tilde{B}_1(t)$ and $\tilde{I}_1(x)$ are represented in figure (\ref{boundary_initial_all_equations}).\\
The derivatives of the Burgers' equation (\ref{burg_pde}) with respect to the variable $x$ and $t$ (corresponding to $x_1$ and $x_2$ in equation (\ref{PDE-residual})) can be evaluated as 
\begin{equation} \label{burg_grad_x}
    \frac{\partial ^2}{\partial x \partial t}(u(t,x)) + \frac{\partial^2}{\partial x^2} (u^2(t,x)/2) = (\eta_v/\pi)\frac{\partial^3}{\partial x^3}(u(t,x))
\end{equation}
and 
\begin{equation} \label{burg_grad_t}
    \frac{\partial^2}{\partial t^2}(u(t,x)) + \frac{\partial^2}{\partial t \partial x}(u^2(t,x)/2) = (\eta_v/\pi) \frac{\partial^2}{\partial x^2}(u(t,x)).
\end{equation}
Subsequently, the loss function for training the neural network can be formulated deploying equation (\ref{loss_with_grad}) from equations (\ref{burg_grad_x}) and (\ref{burg_grad_t}) as
\begin{multline} \label{overall_lossfunc_burger}
    \mathcal{L}_{Burgers'} = \frac{w_G}{|\mathcal{T}_G|} \sum _{(x,t) \in \mathcal{T}_G} \norm{\frac{\partial (\hat{u}(t,x))}{\partial t} + \frac{\partial (\hat{u}^{2}(t,x)/2)}{\partial x} - (\eta_v / \pi) \frac{\partial ^{2}(\hat{u}(t,x))}{\partial x^{2}}}_2^2 \\
    + \frac{w_{D}}{|\mathcal{T}_{D}|}  \sum_{(x,t)\in \mathcal{T}_{D}}\norm{\hat{u}(t,0)-\hat{u}(t,1)}_2^2 + \frac{w_I}{|\mathcal{T}_I|} \sum_{(x,t)\in \mathcal{T}_I}\norm{\hat{u}(0,x)-\sum_{i=1}^{N} \lambda_{i} \sin (2\pi f_{i}/L_x)x - \Psi_{i}}_2^2 \\
    + \frac{w_{g_1}}{|\mathcal{T}_{g_1}|} \sum_{(x,t)\in \mathcal{T}_{g_1}} \norm{\frac{\partial ^2 (\hat{u}(t,x))}{\partial x \partial t} + \frac{\partial^2 (\hat{u}^2(t,x)/2)}{\partial x^2} - (\eta_v/\pi)\frac{\partial^3 (\hat{u}(t,x))}{\partial x^3}}_2 ^2 + \frac{w_{g_2}}{|\mathcal{T}_{g_2}|} \sum_{(x,t)\in \mathcal{T}_{g_2}}\\
    \norm{\frac{\partial^2 (\hat{u}(t,x))}{\partial t^2} + \frac{\partial^2}{\partial t \partial x}(\hat{u}^2(t,x)/2) - (\eta_v/\pi) \frac{\partial^2 (\hat{u}(t,x))}{\partial x^2}}_2^2 + \frac{w_{p}}{|\mathcal{T}_{p}|}  \sum_{(x,t)\in \mathcal{T}_{p}}\norm{\hat{u}(t,x)-\hat{\hat{u}}(t,x)_p}_2^2.
\end{multline}
where $\hat{\hat{u}}_p$ is the pseudo points generated in the previous iterations. If there is supervised dataset available, then an extra loss term $\frac{w_{d}}{|\mathcal{T}_{d}|}  \sum_{(x,t)\in \mathcal{T}_{d}}\norm{\hat{u}(t,x)-u(t,x)_d}_2^2$ can be added to the overall loss function (\ref{overall_lossfunc_burger}).\\
Analyzing the results obtained for the approximated solution $\hat{u}$ of the solution $u$ to the Burgers' equation using the gST-PINN algorithm, we obtain an $L^2$ relative error and $MSE$ error in $\hat{u}$ respectively $7.732166 \times 10^{-2}(\sim <10\%)$ and $5.421603 \times 10^{-5}$, which are exceptional when compared with the results of standard classical PINN, which demonstrates the corresponding values $4.051458 \times 10^{-1}(\sim >10\%)$ and $1.488497 \times 10^{-3}$. Figure (\ref{Burgers_equation_plot_diagram_1}) shows a detailed performance analysis of the gST-PINN model in solving Burgers' equation in terms of predicted solution as a two variable function of space and time variables defined on a rectangular domain, i.e. $\hat{u} ~vs~ (t,x)$ and the absolute error with respect to the true value of the solution from the input labeled data, i.e., $|\hat{u}-u|(x,t)$. The figures not only clearly depict that the solutions estimated by the gST-PINN model are profoundly accurate, but also demonstrates its superiority compared to the standard PINN model in terms of prediction accuracy. Figure (\ref{Burgers_equation_diagram_2}) exhibits the $\hat{u} vs x$ plots at three distinct fixed time values $t=0.3,~0.6,~0.9$. At $t=0.3$, both gST-PINN and PINN provides comparable predictions, nevertheless gST-PINN provides more accurate predictions closely matching the sinusoidal form of solutions. But at higher time values $t=0.6$ and $0.9$, the pre-eminence of gST-PINN becomes conspicuous. While gST-PINN preserves its estimation accuracy over higher time points, it can be seen that the solutions predicted by PINN start to deviate significantly, especially near the extrema as we move to higher time points. For a more precise analysis of accuracy, an inspection of $\hat{u}(t,x)$ from the plots in figure (\ref{Burgers_equation_diagram_2}) shows that $u(0.3,~0.4) \approx 00.1489$, $\hat{u}_{gST-PINN}(0.3,~0.4)\approx 0.1501$, $\hat{u}_{PINN}(0.3,~0.4) \approx 0.1802$. Similar observations can also be made for $t=0.6$ and $t=0.9$.

Both PINN and gST-PINN models implemented fully connected neural networks, structured with 4 neurons of 32 neurons each and the network structure was selected by Bayesian optimization. The network training was done over a spatio-temporal domain spanning $[0,1] \times [0,2]$ discretized into a grid of $N_{x} \times N_{t} = 512 \times 201 = (102, 912 ~~\mathrm{total ~points}).$ The number of initial, boundary and intra-domain data points used were 512, 402 and  respectively. Adam optimizer was used to train both the networks over 20,000 iterations, configured with a learning rate of $10^{-3}$ and the $\tanh$ activation function.\\

\subsection{Diffusion-Reaction Equation} \label{sec_4.2_diff_react}
\begin{figure}[t!]
	\centering
	\hspace*{-1.5cm} 
	\includegraphics[width=6.5in]{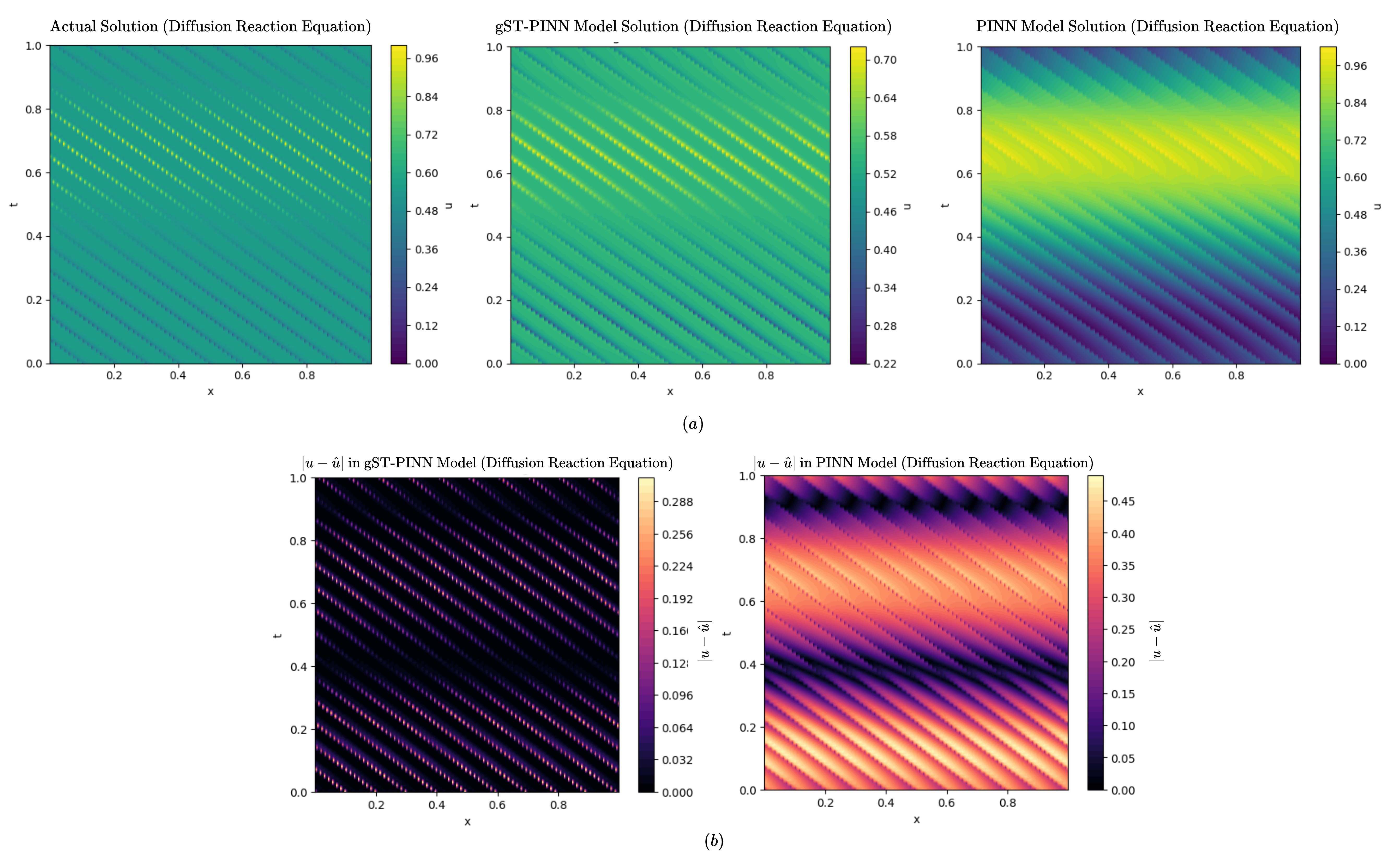}%
	\caption{\textbf{(Section (\ref{sec_4.2_diff_react}), Diffusion-reaction equation)} \textbf{(a)}: Comparison of the actual solution $u$ of diffusion reaction equation with its predicted solutions $\hat{u}$ using gST-PINN and PINN models. \textbf{(b)}: Comparison of the absolute error in $u$, i.e., $|u-\hat{u}|$ predicted by the gST-PINN and PINN model. From both \textbf{(a)} and \textbf{(b)}, it can be clearly summarized that the performance of the gST-PINN model in predicting the solution $u$ of the diffusion reaction equation is substantially better in comparison with the usual PINN model.}
	\label{diff_react_eqn_plot_diagram_1}
\end{figure}

The diffusion-reaction equation in one spatial dimension (also known as Kolmogorov–Petrovsky–Piskunov equation) \cite{zhang2023approximate,wongsaijai2024analytical} in plane geometry can be expressed as given in equation (\ref{diff_react_pde}):
\begin{equation} \label{diff_react_pde}
    \frac{\partial}{\partial t}(u(t,x)) = \xi_v \frac{\partial ^2}{\partial x^2}(u(t,x)) + \rho_m A(u(t,x))
\end{equation}
where $u$ represents the solute concentration. The values of the various parameters in equation (\ref{diff_react_pde}) have been considered by us as\\
$\rho_m$ (\textit{mass density}) $\rightarrow$ 1.5\\
$\xi_v$ (\textit{diffusion coefficient}) $\rightarrow$ 0.5.\\
In this article, we consider a specific diffusion-reaction equation, the Fisher's equation by setting 
\begin{equation}
    A(u(t,x)) = (u(t,x))((1-u(t,x)).
\end{equation}
Similar to the Burger's equation, we consider a periodic boundary condition $(\tilde{B}_{2}(t))$ (\ref{burg_bound_cond}) and the initial condition $(\tilde{I}_{2}(x))$ (\ref{burg_bound_cond}) as a superposition of sinusoidal functions (figure (\ref{boundary_initial_all_equations})).\\
Computing the gradients with respect to the $x$ and $t$ variables, we get the corresponding equations
\begin{equation} \label{diff_react_grad_x}
    \frac{\partial^2}{\partial x \partial t} (u(t,x)) = \xi_v \frac{\partial^3}{\partial x^3}(u(t,x))  + \rho_m \Bigl(\frac{\partial }{\partial x}(u(t,x))\Bigr)(1-2u(t,x))
\end{equation}
and 
\begin{equation} \label{diff_react_grad_t}
    \frac{\partial^2}{\partial t^2} (u(t,x)) = \xi_v \frac{\partial^3}{\partial t \partial x^2}(u(t,x)) + \rho_m \Bigl(\frac{\partial}{\partial t}(u(t,x))\Bigr)(1-2u(t,x)).
\end{equation}
Subsequently, we model the loss function for training the network corresponding to solving the diffusion-reaction equation making use of equations (\ref{diff_react_grad_x}) and (\ref{diff_react_grad_t}) as
\begin{multline}
    \mathcal{L}_{diffusion-reaction} = \frac{w_G}{|\mathcal{T}_G|} \sum _{(x,t) \in \mathcal{T}_G} \norm{\frac{\partial (\hat{u}(t,x))}{\partial t} - \xi_v \frac{\partial ^2 (\hat{u}(t,x))}{\partial x^2} - \rho_m \hat{u}(t,x)(1-\hat{u}(t,x))}_2^2 \\
    + \frac{w_{D}}{|\mathcal{T}_{D}|}  \sum_{(x,t)\in \mathcal{T}_{D}}\norm{\hat{u}(0,t)-\hat{u}(1,t)}_2^2 + \frac{w_I}{|\mathcal{T}_I|} \sum_{(x,t)\in \mathcal{T}_I}\norm{\hat{u}(0,x)-\sum_{i=1}^{N} \chi_{i} \sin (2\pi k_{i}/L_x)x - \Phi_{i}}_2^2 \\
    + \frac{w_{g_1}}{|\mathcal{T}_{g_1}|} \sum_{(x,t)\in \mathcal{T}_{g_1}} \norm{\frac{\partial^2 (\hat{u}(t,x))}{\partial x \partial t}  - \xi_v \frac{\partial^3 (\hat{u}(t,x))}{\partial x^3}  - \rho_m \Bigl(\frac{\partial (\hat{u}(t,x))}{\partial x}\Bigr)(1-2\hat{u}(t,x))}_2 ^2 \\
    + \frac{w_{g_2}}{|\mathcal{T}_{g_2}|} \sum_{(x,t)\in \mathcal{T}_{g_2}}
    \norm{\frac{\partial^2 (\hat{u}(t,x))}{\partial t^2}  - \xi_v \frac{\partial^3 (\hat{u}(t,x))}{\partial t \partial x^2} - \rho_m \Bigl(\frac{\partial (\hat{u}(t,x))}{\partial t}\Bigr)(1-2\hat{u}(t,x))}_2^2\\ 
    \frac{w_{p}}{|\mathcal{T}_{p}|}  \sum_{(x,t)\in \mathcal{T}_{p}}\norm{\hat{u}(t,x)-\hat{\hat{u}}(t,x)_p}_2^2.
\end{multline}
Investigating the results obtained for the estimation of solution $\hat{u}$ of the diffusion reaction equation using gST-PINN model, the $L^2$ relative error and $MSE$ errors in $\hat{u}$ are correspondingly $6.110305 \times 10^{-2}(<6.2\%)$ and $1.067104 \times 10^{-3}$ while the respective values obtained using standar PINN algorithm are $5.218521 \times 10^{-1}(\sim >10\%)$ and $7.783521$, thereby gST-PINN displaying better accuracy. From figure (\ref{diff_react_eqn_plot_diagram_1}), it is clearly visible that the diffusion reaction equation solution estimates are much closer to the actual $u$ values over the entire rectangular $(t,x)$ domain, thus reducing the absolute error over the domain with respect to the actual solution, i.e., $|\hat{u}_{gST-PINN}(t,x) - u(t,x)| \approx \epsilon_1,~|\hat{u}_{PINN}(t,x) -u(t,x)| \approx\epsilon_2 \Rightarrow \epsilon_1 \lesssim \epsilon_2 ~~\forall (t,x)\in \Omega_{diffusion-reaction}$.

\subsection{Diffusion-Sorption Equation} \label{sec_4.3_diff_sorp}
\begin{figure}[h]
	\centering
	\hspace*{-1.7cm} 
	\includegraphics[width=6.5in]{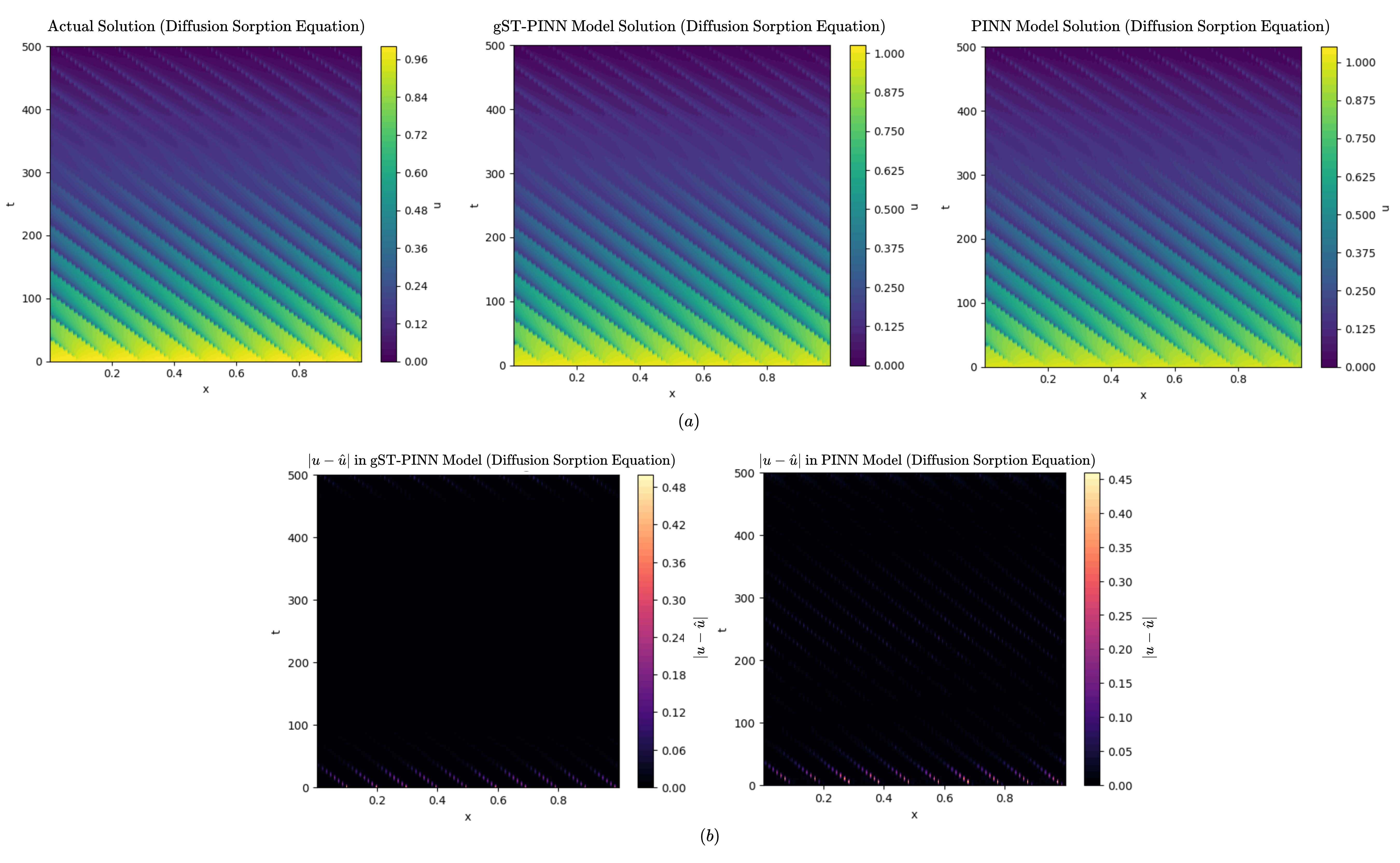}%
	\caption{\textbf{(Section (\ref{sec_4.3_diff_sorp}), Diffusion sorption equation)} \textbf{(a)}: Comparison of the actual solution $u$ of diffusion sorption equation with its predicted solutions $\hat{u}$ using gST-PINN and PINN models. \textbf{(b)}: Comparison of the absolute error in $u$, i.e., $|u-\hat{u}|$ predicted by the gST-PINN and PINN model. From both \textbf{(a)} and \textbf{(b)}, it can be clearly concluded that the gST-PINN model exhibits superior performance in comparison with the usual PINN model for the prediction $\hat{u}$ of the solution $u$ of the diffusion sorption equation.}
	\label{gSTPINN_paper_diffsorption_plots_1}
\end{figure}

\begin{figure}[h]
	\centering
	\hspace*{-1.7cm} 
	\includegraphics[width=6.5in]{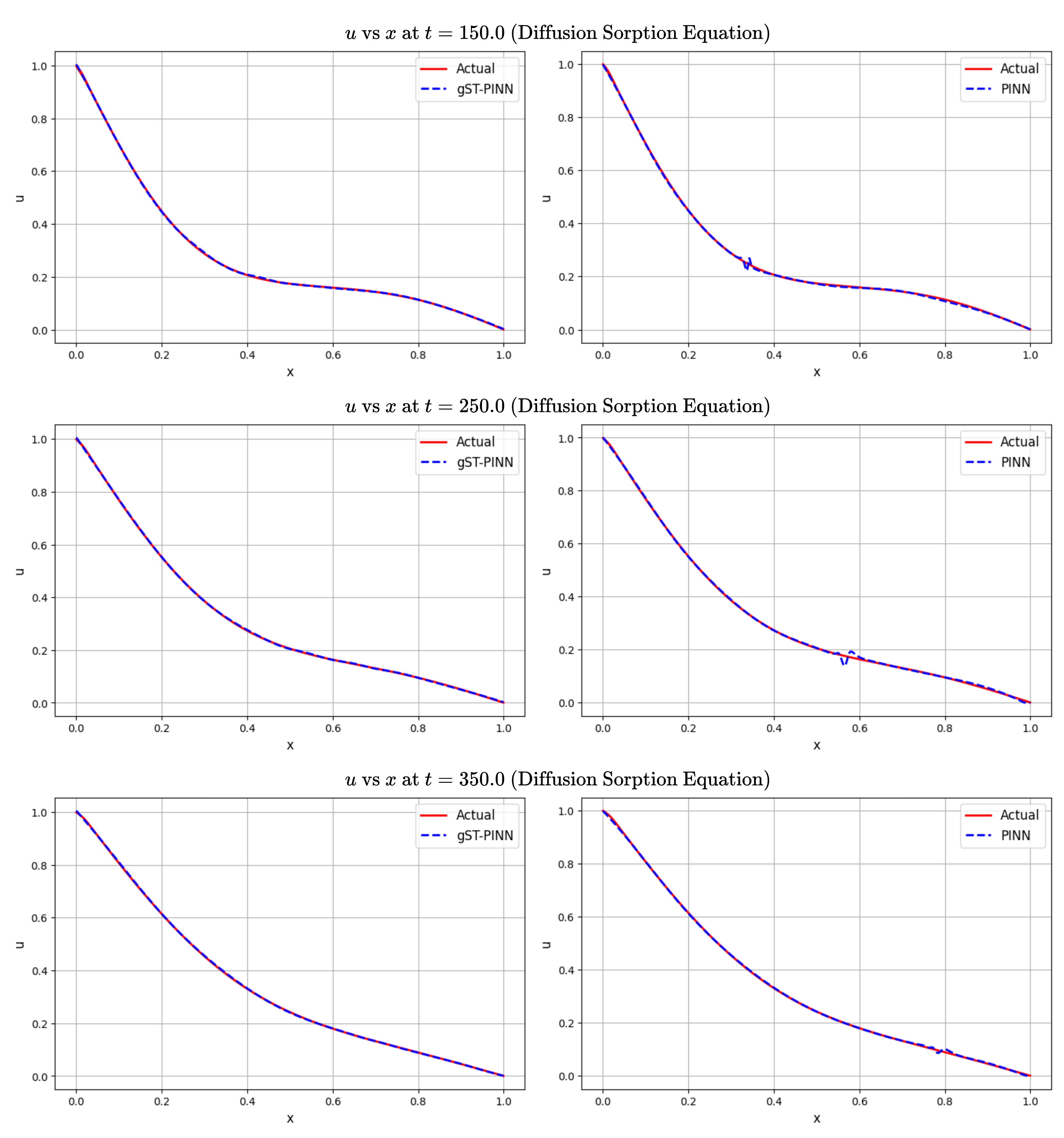}%
	\caption{\textbf{(Section (\ref{sec_4.3_diff_sorp}), Diffusion sorption equation)} "\textbf{u} vs \textbf{x}" plots considered at three distinct time points \textbf{t=150,~250,~350}, of gST-PINN and PINN models for the estimation of solution to diffusion sorption equation. From the plots, it can be clearly examined that PINN portrays lesser accuracy in certain small ranges of $x$ compared to other regions (for example, at $t=150$, $t=250$ and $t=350$ respectively, PINN shows deteriorated performance when $x\in(0.3,~0.35)$, $x\in(0.55,~0.6)$ and $x \in (0.78,~0.81)$) whereas the gST-PINN model portrays exceptional accuracy over the entire range of $x\in [0,1]$ over all the time instances under consideration.}
	\label{gSTPINN_plot_for_diff_sorption_eqn_2}
\end{figure}

The one dimensional diffusion sorption equation can be represented by equation (\ref{diff_sorp_pde}) as 
\begin{equation} \label{diff_sorp_pde}
    \frac{\partial}{\partial t}(u(t,x)) = \frac{C_d}{R(u(t,x))}\frac{\partial^2}{\partial x^2}(u(t,x))~~~~~~x\in (0,1),~t\in(0,500]
\end{equation}
where $R(u(x,t))$ is the retardation factor which is a non-linear function of $u(t,x)$:
\begin{equation} \label{diff_sorp_ret_factor}
    R(u(t,x)) = 1 + \frac{1-\psi}{\psi}\rho_{s}k_{f}n_{f}u^{n_{f}-1}.
\end{equation}
The values of the various parameters in equations (\ref{diff_sorp_pde}) and (\ref{diff_sorp_ret_factor}) has been considered to be\\
$\psi$ (\textit{porosity} of the medium for which the PDE is defined) $\rightarrow$ $0.31$\\
$k_f$ (\textit{Freundlich's parameter}) $\rightarrow$ $3.5 \times 10^{-4}$\\
$n_f$ (\textit{Freundlich's exponent}) $\rightarrow$ $0.875$\\
$\rho_s$ (\textit{bulk density}) $\rightarrow$ $2875$\\
$C_s$ (\textit{diffusion coefficient}) $\rightarrow$ $4.5 \times 10^{-4}$\\
At the lower boundary $(x=0)$, we have the uniform Dirichlet boundary condition given by 
\begin{equation} \label{diff_sorp_bound_1}
    u(t,0) = 1 \approx \tilde{B}_3(t)
\end{equation}
and at the higher boundary, we have a Robin boundary condition:
\begin{equation} \label{diff_sorp_bound_2}
    u(t,1) = C_s\frac{\partial u}{\partial x}(t,1).
\end{equation}
The initial condition is formulated using a uniform random distribution:
\begin{equation} \label{diff_sorp_ini}
    u(0,x)=0.16395~~~\forall x\in (0,1) \approx \tilde{I}_3(x).
\end{equation}
The plot of initial and boundary conditions are displayed in figure (\ref{boundary_initial_all_equations}).\\
The derivatives (or gradients) with respect to the space and time variables can be computed respectively as 
\begin{multline} \label{diff_sorp_grad_x}
    \frac{\partial ^2}{\partial x\partial t} u(t,x) = \frac{C_d}{R(u(t,x))} \Bigl\{\frac{\partial^3 }{\partial x^3}u(t,x) - \Bigl(\frac{1}{R}\Bigr) \Bigl(\frac{\partial}{\partial x}u(t,x)\Bigr) \Bigl(\frac{\partial^2 }{\partial x^2}u(t,x)\Bigr) \Bigl(\frac{1-\psi}{\psi}\Bigr)\\(\rho_s n_f k_f (n_f-1))u^{n_f-2}\Bigr\}
\end{multline}
and 
\begin{multline} \label{diff_sorp_grad_t}
    \frac{\partial^2 }{\partial t^2}u(t,x) = \frac{C_d}{R(u(t,x))} \Bigl\{\frac{\partial ^3}{\partial t \partial x^2} u(t,x) -\Bigl(\frac{1}{R}\Bigr)\Bigl(\frac{\partial}{\partial t}u(t,x) \Bigr) \Bigl(\frac{\partial^2}{\partial x^2}u(t,x)\Bigr) \Bigl(\frac{1-\psi}{\psi}\Bigr)\\(\rho_s n_f k_f (n_f-1))u^{n_f-2}\Bigr\}.  
\end{multline}
Hence, from equations (\ref{diff_sorp_pde}), (\ref{diff_sorp_bound_1}), (\ref{diff_sorp_bound_2}), (\ref{diff_sorp_ini}), (\ref{diff_sorp_grad_x}) and (\ref{diff_sorp_grad_t}), the loss function for training the network can be constructed as 
\begin{multline}
    \mathcal{L}_{diffusion-sorption} = \frac{w_G}{|\mathcal{T}_G|} \sum _{(x,t) \in \mathcal{T}_G} \norm{\frac{\partial (\hat{u}(t,x))}{\partial t} - \frac{C_d}{R(\hat{u}(t,x))}\frac{\partial^2 (\hat{u}(t,x))}{\partial x^2}}_2^2 \\
    + \frac{w_{D_1}}{|\mathcal{T}_{D_1}|}  \sum_{(x,t)\in \mathcal{T}_{D_1}}\norm{\hat{u}(t,1) - C_d\frac{\partial \hat{u}}{\partial x}(t,1)}_2^2 + \frac{w_{D_2}}{|\mathcal{T}_{D_2}|}  \sum_{(x,t)\in \mathcal{T}_{D_2}}\norm{\hat{u}(t,0) - 1}_2^2 + \frac{w_I}{|\mathcal{T}_I|} \\
    \sum_{(x,t)\in \mathcal{T}_I}\norm{\hat{u}(0,x)-0.16395}_2^2 + \frac{w_{g_1}}{|\mathcal{T}_{g_1}|} \sum_{(x,t)\in \mathcal{T}_{g_1}} \bigg|\bigg|\frac{\partial ^2}{\partial x\partial t} \hat{u}(t,x) - \frac{C_d}{R(\hat{u}(t,x))} \Bigl\{\frac{\partial^3 \hat{u}(t,x)}{\partial x^3}\\
    - \Bigl(\frac{1}{R}\Bigr) \Bigl(\frac{\partial \hat{u}(t,x)}{\partial x}\Bigr) \Bigl(\frac{\partial^2 \hat{u}(t,x)}{\partial x^2}\Bigr) \Bigl(\frac{1-\psi}{\psi}\Bigr)(\rho_s n_f k_f (n_f-1))\hat{u}^{n_f-2}\Bigr\}\bigg|\bigg|_2 ^2 + \frac{w_{g_2}}{|\mathcal{T}_{g_2}|} \sum_{(x,t)\in \mathcal{T}_{g_2}}\\
    \bigg|\bigg|\frac{\partial^2 }{\partial t^2}\hat{u}(t,x) - \frac{C_d}{R(u(t,x))} \Bigl\{\frac{\partial ^3}{\partial t \partial x^2} \hat{u}(t,x) -\Bigl(\frac{1}{R}\Bigr)\Bigl(\frac{\partial}{\partial t}\hat{u}(t,x) \Bigr) \Bigl(\frac{\partial^2}{\partial x^2}\hat{u}(t,x)\Bigr) \Bigl(\frac{1-\psi}{\psi}\Bigr)\\
    (\rho_s n_f k_f (n_f-1))\hat{u}^{n_f-2}\Bigr\}\bigg|\bigg|_2^2 + \frac{w_{p}}{|\mathcal{T}_{p}|}  \sum_{(x,t)\in \mathcal{T}_{p}}\norm{\hat{u}(t,x)-\hat{\hat{u}}(t,x)_p}_2^2.
\end{multline}
When considering the diffusion-sorption equation (\ref{diff_sorp_pde}), for the predicted solution $\hat{u}$, the gST-PINN model yields $L^2$ relative error and $MSE$ errors respectively $1.831367 \times 10^{-2} (\sim <2\%)$ and $5.348408 \times 10^{-5}$. Alike for the case of Burgers' equation, the gST-PINN model shows superior performance also for diffusion-sorption equation compared to PINNs which yields the $L^2$ relative errors and $MSE$ errors of $2.330058 \times 10^{-2} (\sim >2\%)$ and $8.657794 \times 10^{-5}$ respectively for the predicted solution $\hat{u}$. Further performance analysis of the gST-PINN and its comparison with the standard PINN model can be examined from the figures (\ref{gSTPINN_paper_diffsorption_plots_1}) and (\ref{gSTPINN_plot_for_diff_sorption_eqn_2}), which respectively are the $\hat{u} ~vs~(x,t)$ information across the 2-D rectangular domain and prediction of the form of solution as a function of $x$ at three distinct fixed times $t=150,~250,~350$, i.e., $\hat{u}(x,150),~\hat{u}(x,250),~\hat{u}(x,350)$.     

\subsection{Comparison with other scientific machine learning models} \label{sec_4.4_comp_oth_models}
\begin{figure}[t!]
	\centering
	\includegraphics[width=5.0in]{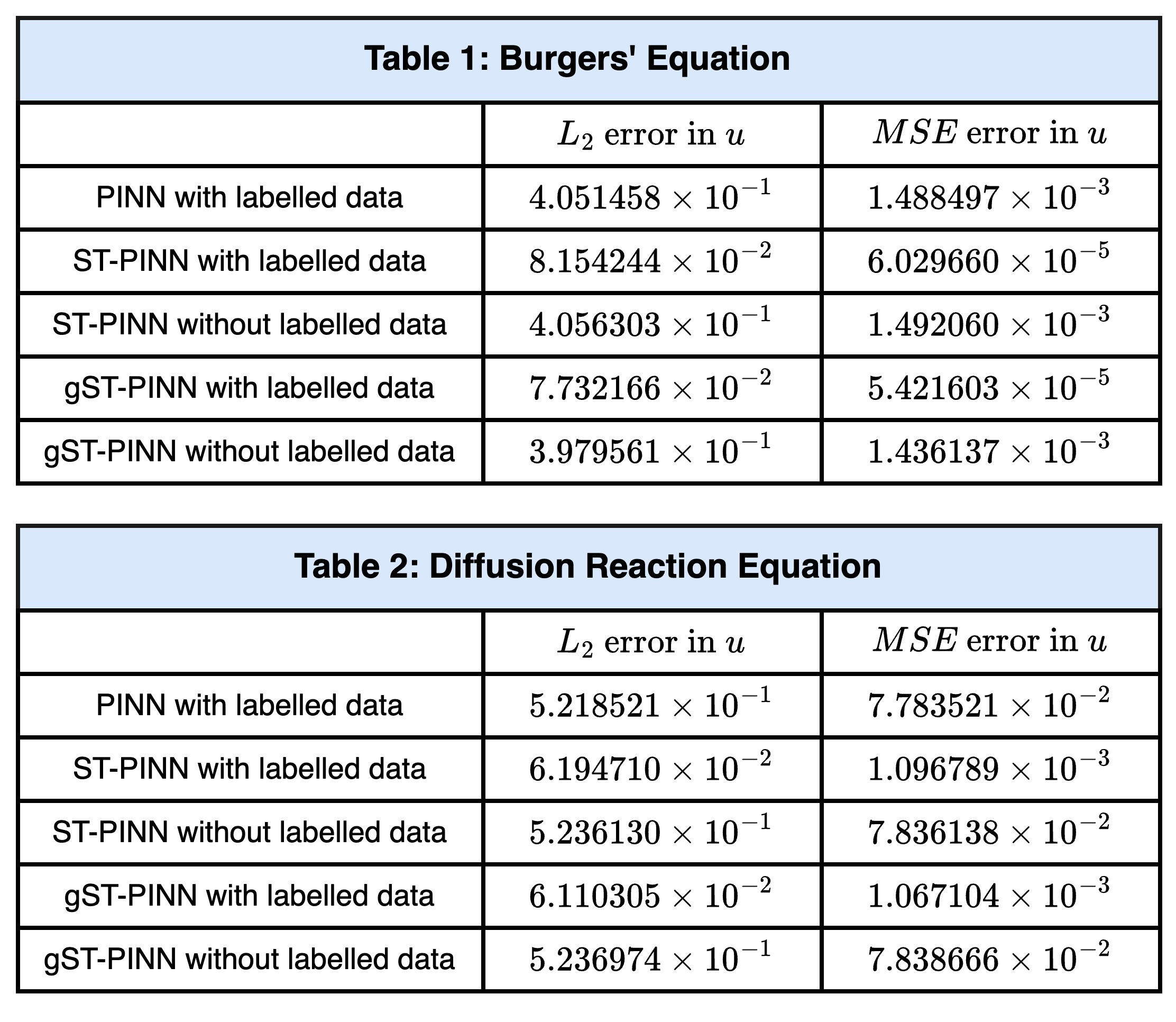}%
	\caption{\textbf{Section (\ref{sec_4.4_comp_oth_models})} Comparison of gST-PINN \textbf{(Table 1: Burgers' equation, Table 2: diffusion reaction equation)} with some state-of-the-art scientific machine learning models. The gST-PINN model with the incorporation of a few labelled data outperforms all the other models (PINN, gPINN, ST-PINN) in terms of accuracy in the prediction of solution to nonlinear PDEs.}
	\label{figure_comp_table}
\end{figure}

For portraying the dominance of the novel gST-PINN model over other existing scientific machine learning models for solving non-linear partial differential equations, we provide a performance comparison table which depicts the performance of various state-of-the-art models in terms of $L_2$ error and $MSE$ error in the solution $u$ for solving Burgers' and Diffusion Reaction equations. For our analysis, we consider the models PINN (Physics Informed Neural Network), ST-PINN (Self Training Physics Informed Neural Network) \cite{yan2023st} and our newly proposed gST-PINN (Gradient Enhanced Self Training Physics Informed Neural Network). For both the equations, in accordance with the trivial expectations, augmentation of a few labeled data to the input enhances the model efficiency for all the models. For the Burgers' equation, PINN with labeled dataset gives an $L_2$ and $MSE$ error of $4.051458 \times 10^{-1}$ and $1.488497 \times 10^{-3}$ respectively. ST-PINN without labeled data also shows approximately similar results with negligible improvement with the values $4.056303 \times 10^{-1}$ and $1.492060 \times 10^{-3}$ respectively. ST-PINN with the addition of a few labeled dataset in the input shows substantial improvement in performance metrics with corresponding $L_2$ and $MSE$ erros in $u$ $8.154244 \times 10^{-2}$ and $6.029660 \times 10^{-5}$. Now, analysing the figures of the gST-PINN model, it exhibits profound escalation in performance when compared with the other two models. gST-PINN without labeled data shows better results than both PINN and ST-PINN without labeled data with the value of $L_2$ and $MSE$ errors in $u$ equal to $3.979561 \times 10^{-1}$ and $1.436137 \times 10^{-3}$. gST-PINN with labeled data depicts the best accuracy among all the models considered with comparatively exceptional figures of $7.732166 \times 10^{-2}$ and $5.421603 \times 10^{-5}$. For the diffusion reaction equation also, the trend is similar to Burgers' equation as depicted by table (\ref{figure_comp_table}). PINN with the inclusion of labeled data incorporates $L_2$ and $MSE$ errors in $u$ respectively $5.218521 \times 10^{-1}$ and $7.783521 \times 10^{-2}$ where as ST-PINN model with and without labeled data shows corresponding errors $6.194710 \times 10^{-2}$, $1.096789 \times 10^{-3}$, $5.236130 \times 10^{-1}$ and $7.836138 \times 10^{-2}$. The gST-PINN model without labeled data shows $L_2$ and $MSE$ errors in $u$ $5.236974 \times 10^{-1}$ and $7.838666 \times 10^{-2}$ while the analogous values for gST-PINN model inclusive of labeled data are $6.110305 \times 10^{-2}$ and $1.067104 \times 10^{-3}$.\\
All these above mentioned statistics indicate the supremacy of gST-PINN model for approximating the solution of highly non linear PDEs, in terms of both accuracy and efficiency in comparison with other models. It is noteworthy that Some of the models like the gPINN \cite{yu2022gradient}, gw-PINN \cite{xiong2023gradient}, PINN without labeled data etc. have not been considered for the comparisons as these models exhibit substantially low results with respect to the models taken into consideration for the analysis.

\section{Conclusion and future outlook}
In the current research article, we have proposed a novel gradient enhanced self training physics informed neural network (gST-PINN) model, which is based on the semi-supervised learning paradigm. The core functioning of the model is based on our formulated gradient enhanced pseudo labeling algorithm which utilizes the information on the PDE residual and its derivative with respect to each independent variable to generate supervised data from unsupervised collocation points (gradient enhanced pseudo labeling) upon satisfying certain threshold filtration criteria. Furthermore, the algorithm is characterized by two crucial hyperparameters, the pseudo label selection threshold $r$ and the psuedo label candidate rate $q$. We have utilized the gST-PINN model to solve some of the most important non-linear PDEs including the Burgers' equation, diffusion-reaction equation and diffusion-sorption equation to demonstrate the superiority of gST-PINN model compared to the classical PINN in terms of accuracy in estimating the solution. For respectively the Burgers' equation, diffusion-reaction equation and diffusion-equation, the gST-PINN model achieves a $MSE$ error in the approximated solution $\hat{u}$ equal to $5.421603 \times 10^{-5}$, $1.067104 \times 10^{-3}$ and $5.348408 \times 10^{-5}$ compared to the figures $1.488497 \times 10^{-3}$, $7.783521 \times 10^{-2}$ and $8.657794 \times 10^{-5}$ achieved by the classical PINN model. Subsequently, it has also been demonstrated that the performance of the gST-PINN model without the inclusion of the labeled input data (with $1.436137 \times 10^{-3}$ and $7.838666 \times 10^{-2}$ being the $MSE$ errors in $\hat{u}$ for Burgers' equation and diffusion reaction equation respectively) is comparable to that of the performance portrayed by the classical PINN model assisted with inclusion of supervised input data. Furthermore, we also show the superiority of our proposed gST-PINN model with the inclusion of a small amount of supervised input data over the state-of-the-art models including the classical PINN, ST-PINN and gPINN, all considered with the provision of labeled input data.\\
One of the primary scopes of for future research as an extension to this research work include the modification of the gST-PINN model to address the geometry and topology of the PDE domain in order to solve high dimentional PDEs defined on non-trivial domains (such as moving boundaries) and Riemannian manifolds. Another noteworthy scope for future research is the enhancement of the gST-PINN model by augmenting it with the ability to solve fractional differential equations, as currently the number of scientific machine learning models specifically designed to exhibit high accuracy in solving fractional differential equations is scarce.









\bibliography{sn-bibliography}

\end{document}